# Design and Development of a Robotic Vehicle for Shallow-Water Marine Inspections

**Parag Tarwadi[1], Yuta Shiraki[2], Ori Ganoni[3], Shanghai Wei[4], Ho Seok Ahn[5], Bruce MacDonald[6]**

[1,5,6]Centre for Automation and Robotic Engineering Science, University of Auckland, Auckland, New Zealand,
[1]ptar501@aucklanduni.ac.nz, [5,6]{hs.ahn; b.macdonald}@auckland.ac.nz
[2]School of Science and Engineering, Ritsumeikan University, Shiga, Japan
u.ta.s.0921@gmail.com
[3]Computer Graphics and Image Processing Group, University of Canterbury, Christchurch, New Zealand
ori.ganoni@pg.canterbury.ac.nz
[4]Department of Chemical and Materials Engineering, University of Auckland, Auckland, New Zealand,
s.wei@auckland.ac.nz

## Abstract

Underwater marine inspections for ship hull or marine debris, etc. are one of the vital measures carried out to ensure the safety of marine structures and underwater species. This work details the design, development and qualification of a compact and economical observation class Remotely Operated Vehicle (ROV) prototype, intended for carrying out scientific research in shallow-waters. The ROV has a real-time processor and controller onboard, which synchronizes the movement of the vehicle based on the commands from the surface station. The vehicle piloting is done using the onboard Raspberry pi camera and the support of some navigation sensors like Global Positioning System (GPS), inertial, temperature, depth and pressure. This prototype of ROV is a compact unit built using a limited number of components and is suitable for underwater inspection using a single camera. The developed ROV is initially tested in a pool.

## 1 Introduction

The underwater inspection for the marine debris is essential for ensuring the marine environmental safety. Solid waste in the sea or river is known as marine debris. Marine debris has been recognized as a form of pollution for nearly 50 years and is a concern for human health and safety [Ryan, 2015]. A more recently discovered problem with marine debris is mainly plastic resins, which is the way of introduction of toxic chemicals in the marine environment [Gall et al., 2015]. The underwater inspection of the ship hull is also one of the vital and routine tasks from the viewpoint of fuel consumption and ecosystem conservation [Ishizu et al., 2012]. Several sessile organisms found on ship hulls are invasive and affect native species, resulting in changes to the ecosystem [Kawamura, 2015]. The general underwater marine inspection is mostly carried out by divers. However, this method is not suitable for a long time working, and the divers could be exposed to hazardous situations [Wei et al., 2013]. The underwater inspections by using ROV can be an advantageous alternative. ROV is a teleoperated vehicle for underwater applications [Govindarajan et al., 2013]. The primary goals of this paper are to design and develop an ROV for scientific research work which,

- has light in weight,
- provides an effective and reliable control system with required sensors and camera,
- is neutrally buoyant and stable underwater,
- provide speed up to 2 knots (1.02 m/s),
- is able to carry out shallow water inspection tasks efficiently for marine debris and ship hull.

For these purposes, we developed a specialized ROV prototype, which is as shown in Figure 1,

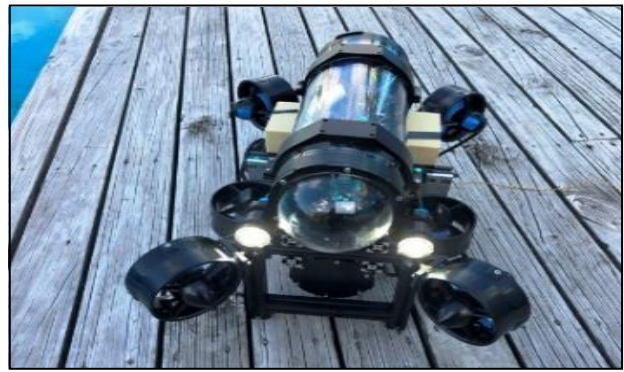

Figure 1. Real view of the developed ROV.

Table 1 below provides the detailed specifications of the developed ROV.

Table 1. Specifications of the developed ROV

| | |
|---|---|
| **Material of Frame** | Black anodized aluminum alloy |
| **Weight** | 8.6 Kg (without Foam) <br> 9.24 Kg (With Foam) |
| **Dimensions** | 0.54 m x 0.34 m x 0.31 m |
| **Diving Depth** | Up to 10 m |
| **Average Speed** | 1.92 Knots (0.99 m/s) |
| **Buoyancy** | Subsea polyurethane foam (0.68 Kg) |
| **Power** | 14.8 VDC, 0.182 kW |
| **Thrusters** | Six Blue Robotics T-100 thrusters |
| **Cable** | 2 wire Cable of 90 m |
| **Data Telemetry** | Normal two-wire cable |
| **Hardware** | Raspberry pi 3B, Arduino Uno |
| **Software** | Qground Control, Putty |
| **Camera** | Raspberry pi camera |
| **Lights** | 2 lumen subsea lights |
| **Navigation Sensors** | LS20031 GPS receiver, 30 bar pressure/ depth, Current and Voltage Sensing and inertial MPU-9250 (3-DOF Gyroscope, 3-DOF Accelerometer, 3-DOF Magnetometer) sensor |

ROVs can be categorized into two major classes based on the purpose of use and their functions, which are observation class ROVs and work class ROVs. Observation class ROVs are generally used for visual inspection and light intervention tasks, whereas work class ROVs perform more serious deep-water tasks with wide power variations [Capocci et al., 2017].

Our developed ROV is an observation or inspection class mini ROV, which can perform shallow-water marine inspections tasks for the ship hull and marine debris etc. This paper is divided into five sections. The first section explains the necessity for the underwater inspection with the introduction of the ROV system. The second section is about the previous related works. The third section will be the design and development of the prototype and the fourth section is about the field trials of the developed ROV with discussion. Finally, the conclusion section, which reiterates the main contributions of the work and highlights some of the possible future improvements.

## 2 Related Works

[Capocci et al., 2017] presents a review about the classification of ROVs in terms of their size and capability and discusses common subsystems of the ROV. Generally, observation class ROVs also called inspection ROVs, are small vehicles deployed in waters no deeper than a few meters and their propulsion power is limited to several kilowatts. This class of ROVs can be subdivided into micro, mini and medium ROVs according to the size of the vehicle. They are often fitted with thrusters, imaging devices and various types of sensors. Work class ROVs can be divided into light and heavy work class models based on the level of heavy-duty work they can carry out. However, work class ROVs employ a considerable volume of equipment that leads to high overall system complexity and high costs for operation. Hence, when the functionalities of these large ROVs are not required, observation class ROVs are preferred for a variety of applications [Christ et al., 2007].

[Sandøy, 2016] designed a model-based advanced control system containing both an observer and a controller for a mini ROV called Blue ROV from Blue Robotics. The controller uses the estimated states produced by the observer and evaluates optimized corrective signals to control the vehicle. Although the Blue ROV includes six thrusters, the system was simplified to the four Degrees of Freedom (DOF), which provides the movement control of surge, sway, heave and yaw motions. The author validated the design of the system by the implementation in Simulink and interfaced with the Robot Operating System (ROS). [Aili et al., 2016] also modeled and developed a control system for Blue ROV in which the model parameters were estimated using EKF-based sensor fusion method to design attitude, angular velocity and depth controllers. However, the attitude controller was not able to achieve a stable system while using the feedback linearization.

Image capturing is another main capability for ROVs. [Jakobsen, 2011] developed a software system for a micro ROV to inspect fish cage net integrity by analyzing the video feed from the ROV. The developed algorithm processes visual data in real-time to generate control signals for the ROV to scan the whole cage net. Although it was not achieved as a result of the lack of sway motion ability of the ROV, but has proven the use of an autonomous ROV for aquaculture monitoring applications. [Aras et al., 2009] were designed a cylindrical shaped underwater vehicle with the use of propellers and a water pump mechanism for the movement. When water inserts in the water tank of a vehicle, it was moving downward, and when the water injected out of the water tank, the vehicle moves upward. [Ishizu et al., 2012] presented a small underwater robot with a mechanical contact mechanism, a hand-eye vision system, and a stereo camera system. This system needs one operator to control a commercial underwater robot and an additional operator to control the hand-eye vision system. This system is suitable only for small underwater robots while considering a steady flow for numerical analysis. [Singh et al., 2015] were constructed a 4 DOF hull shaped ROV. They found that cylindrical hull ROV was better in performance than spherical hull ROV. [Subramanian et al., 2014] were developed a surveillance robot, which was transmitting data such as live video and pictures in the marine environment. They tested an underwater robot with a 4m depth. In oceanography research, there are application examples in detecting and tracking underwater objects, underwater environment mapping and reconstruction [Walther et al., 2004; Marsh et al., 2013].

## 3 System Design

### 3.1 Mechanical Structure

The mechanical design of the ROV was made in a symmetrical way to make a construction relatively straightforward and to simplify software modelling. Based on the characteristic comparison between various materials previously used for the construction of ROV shown in Figure 2, we chose black anodized aluminium alloy as a material to make the frame of an underwater vehicle. While considering the development of mobile robots, aluminium and steel are the most common metals. Aluminium is a softer metal and is, therefore, easier to work with, but steel is stronger. Aluminium is common, cost-effective and easy to assemble. It is lightweight as well as has an excellent ability to resist corrosion, which would help to protect the frame from the harsh salt water and chlorine environment [azom, 2019].

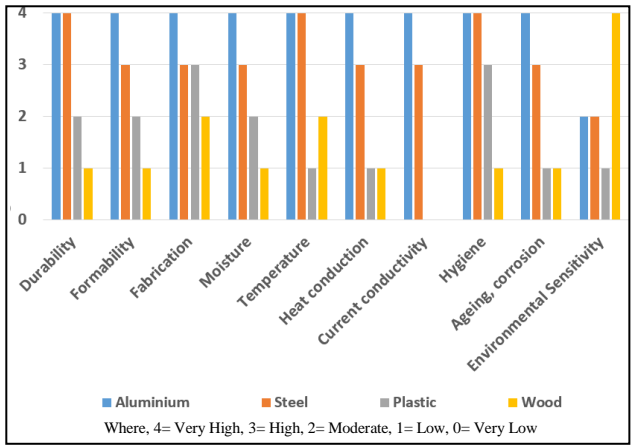

Figure 2. Material characteristic comparison [azom, 2019].

Figure 3 represents the complete mechanical architecture of the ROV by using Solid Edge. The developed vehicle is an open frame mini observation or inspection class ROV, which is equipped with six thrusters. The vehicle comprises two watertight enclosures machined from acrylic separately mounted to a supporting aluminium frame. The first enclosure contains the control system and camera, while another includes a battery. Two high-intensity luminous lights attached at the front end of the ROV for clearer visual observations. The vehicle weighing 8.6 kg in the air is made slightly positively buoyant in water, by using subsea polyurethane foam mounted on both sides of the vehicle.

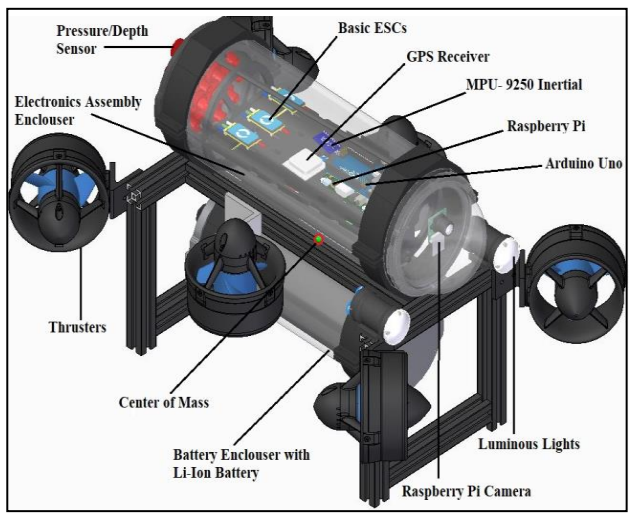

Figure 3. Design of the ROV by using Solid Edge.

### 3.2 Electronic System

The developed electronic model of the ROV is cost-effective, reliable and efficient to control the movements of the vehicle. Figure 4 represents a schematic diagram of the electronic assembly of the ROV, which was controlled by using a surface station (laptop) and joystick. This surface station is connected to a two-wire converter through LAN cable. The LAN is connected to the electronic circuitry part of the ROV.

Raspberry Pi 3 B is used as a mediator between the user interface and the control unit of the ROV. It will send movement controlling actions, which will be received from the surface station to the Arduino. Also, it will send pressure and 9-axis sensor outputs to the user interface. The control unit includes a single Arduino Uno board, electronic speed controllers (ESC) and thrusters. The Arduino Uno is responsible for controlling the speed and movement of thrusters through ESCs. For this, PWM signal pins of Arduino Uno were used. Also, Arduino UNO controls the brightness of the lights. A specially designed battery is used for this work. The thrusters, lights, and ESCs are directly working on this power supply, while for sensors, Arduino Uno and Raspberry Pi 3B a different power source is required, which was achieved by using 5 V and 3 V power converters [Yue et al., 2013].

The camera is an eye of the underwater robotic vehicle. Here we are using the Raspberry Pi camera, which is directly connected to the Raspberry Pi 3B. The surface station uses the Secure Socket Shell (SSH) protocol to communicate with Raspberry Pi 3B by using Putty software [Williams et al., 2001]. It uses the User Datagram Protocol (UDP) socket with the help of joystick to send control signals to the electronic assembly [Nahon, 1996]. Sensors and Arduino Uno were connected to the Raspberry Pi 3B by using I2C and serial communication, respectively.

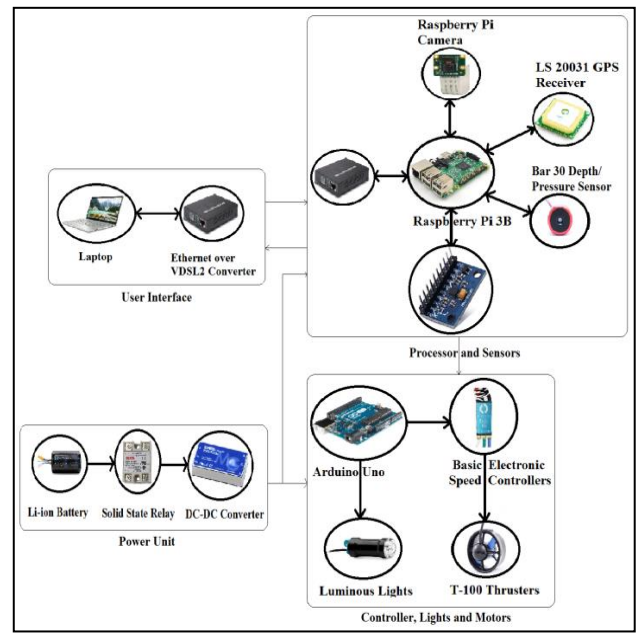

Figure 4. ROV system diagram.

## 3.3 Simulation

To create a simulation environment for the indoor testing of ROV design, a configuration was used that includes the DroneSimLab framework [Ganoni et al., 2017; 2019] and a 3D solid edge model of the ROV. This simulation is used for future indoor testing of the ROV. Figure 5 represents a simulation environment with the developed ROV, and a real-time video of the camera output based on the position and attitude information of the vehicle.

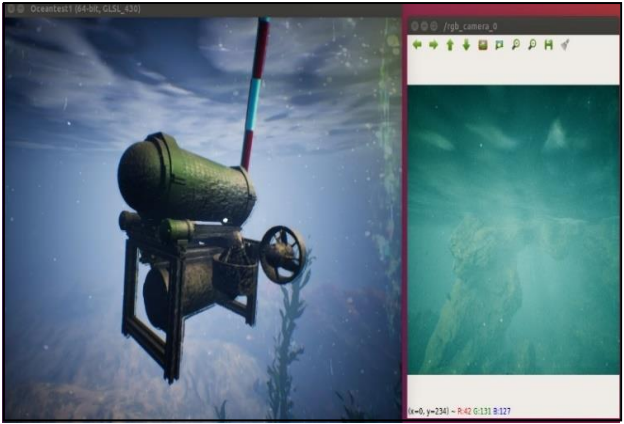

Figure 5. ROV in a simulated environment and the camera output.

## 3.4 Stability Analysis

Stability of the underwater vehicle is affected by the distance between the center of gravity ($Cg$) and the center of buoyancy ($Cb$). To ensure the stability of the vehicle, it is required that the $Cb$ is higher than $Cg$ [Fossen, 2011]. The vehicle will float or sink depending upon the net effect of the weight of the vehicle and the buoyant force generated by the vehicle [Chotikarn et al., 2010]. Based on this, there are three possible conditions for the object as shown in below table 2,

Table 2. Buoyancy conditions [Blidberg, 2001]

| Case | Condition | Result |
| --- | --- | --- |
| buoyant force > Weight | Positive buoyancy | Vehicle floats |
| buoyant force = Weight | Neutral buoyancy | Vehicle neither floats or sinks |
| buoyant force < Weight | Negative buoyancy | Vehicle sinks |

Neutral buoyancy does not allow the vehicle to float or sink while underwater [Nahon, 1996]. So, we made the buoyancy of the vehicle slightly positive; this allows the vehicle to float back to the surface at a very slow rate. To achieve this, we need to calculate the net buoyancy. Before that, first, we need to calculate, $Fg$ (the function of the mass of the object) and $Fb$ (the function of the mass of the water displaced). As the mass of the vehicle is 8.6 kg and the displacement of the vehicle is 7918827.87 mm³, so based on,

$$F = mg. \quad (1)$$

$Fg$ is 84.28 N and $Fb$ is 77.60 N. Therefore, it is observed that the $Fg$ is more than $Fb$ and as a result, the vehicle will sink. So, to achieve the stability of the vehicle, we need to calculate the net buoyancy,

$$Net\ Buoyancy = Fb - Fg \quad (2)$$
$$= -0.68\ kg.$$

Based on equation (2), we added 0.68 kg of subsea polyurethane foam on the vehicle. As an output, now the updated mass of the vehicle is 9.24 kg and the displacement of the vehicle is 9993692.37 mm³. Table 3 below explains the $Cg$ and $Cb$ values for the developed ROV after adding foam with the representation of it as shown in Figure (6),

Table 3. $Cg$ and $Cb$ values

| | | |
| --- | --- | --- |
| **Centre of Gravity ($Cg$)** | $Xcg$ | 0.20 mm |
| | $Ycg$ | 127.47 mm |
| | $Zcg$ | -100.81 mm |
| **Centre of Buoyancy ($Cb$)** | $Xcb$ | 0.29 mm |
| | $Ycb$ | 130.32 mm |
| | $Zcb$ | -98.34 mm |

So, based on equation (1), (2), table 3 and Figure 6, the updated $Fb$ (97.93 N) is more than $Fg$ (90.55 N) and $Cb$ is higher than $Cg$; as a result, the vehicle is slightly positively buoyant.

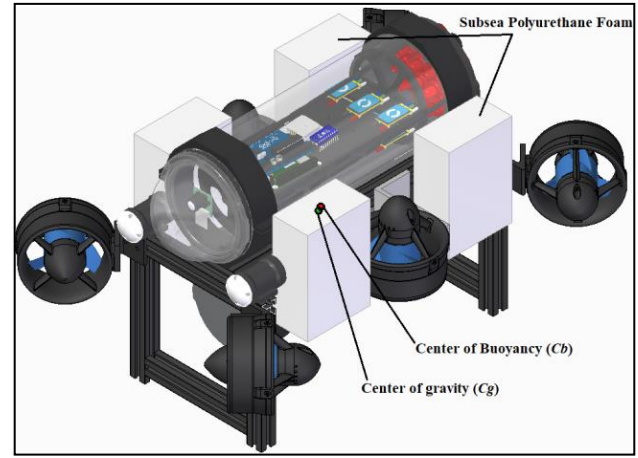

Figure 6. Stability analysis of the developed ROV.

The ROV control and motion highly depend on the thruster configuration. DOF is one of the crucial factors that affect the thruster configuration. DOF provides us with every possible movement of the vehicle. These movements include rotations and displacement along x, y, and z Cartesian axes [Craven et al., 1998; Stutters et al., 2008]. As shown in the following Figure 7, our developed ROV uses six thruster configuration.

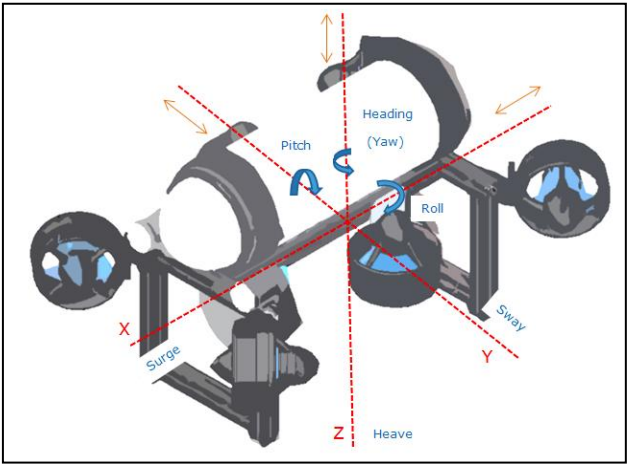

Figure 7. Possible orientations for the developed ROV.

The two thrusters are used for vertical motion, while the remaining four thrusters are used for horizontal motion and positioned at 45 degrees from each other for more DOF. These adjustments allowing vehicle to heave (movement along the vertical plane), surge (longitudinal travel along a horizontal plane), sway (lateral movement along a horizontal plane), roll (rotation about the longitudinal axis (x)), pitch (rotation about the lateral axis (y)) and heading (rotation about the vertical axis (z)) motions. The chosen thruster configuration allows for a more intuitive control for the pilot and a more powerful horizontal movement.

## 4 Field Trials in a Pool and Discussion

The initial testing trials of the developed ROV are carried out in the pool, which for vehicle stability, buoyancy and reliable performance of electronic and electrical systems. The vehicle was tested for the design performance in the pool of the BoxFish research company during early May 2019, wherein the vehicle was tested up to a water depth of 5m. To confirm the stability of the vehicle, we shut the power and then submerged it in the pool without adding foam. The ROV reaches to the ground level of the pool in 8.43 seconds. It means that the ROV is negatively buoyant. To make the ROV stable with slightly positive buoyancy, we again submerged the ROV in the pool with added subsea polyurethane foam at the top. As a result, ROV goes down in the pool till 1.1 m of depth in 2.75 seconds and comes back to the surface water in 3.2 seconds. During the tests, the subsystems of the vehicle were tested for the desired functionality.

### 4.1 Trial using Waves

To see the general behavior of the ROV at various depths, we initially submerged the ROV in pool water. The experiment was conducted by generating waves while considering their height (H) and length (L), as shown in table 4. Let a heading angle against coming waves indicate 0 degrees, the clockwise direction of rotation. As per trial number 3, the horizontal rotation is not started for the big waves when the submerged depth is small. On the other hand, as per trial numbers 4 and 5, the horizontal rotation is not generated for small waves when the submerged depth increases to some extent. According to the trial numbers 7, 8 and 9, the horizontal rotation for ROV is controlled as the submerged depth of the ROV is 3-4 times more than the actual height of the ROV.

Table 4. Waves and submerged depth variations

| No. | Depth (m) | Wave (m) H | Wave (m) L | Initial Heading (deg) | Horizontal Heading (deg) |
|---|---|---|---|---|---|
| 1 | 0.5 | 0.10 | 1.44 | 0 | -90 |
| 2 | 0.5 | 0.20 | 2.21 | -90 | 0 |
| 3 | 0.5 | 0.20 | 2.21 | 0 | 0 |
| 4 | 0.8 | 0.10 | 1.44 | 0 | 0 |
| 5 | 0.8 | 0.15 | 1.75 | 0 | 0 |
| 6 | 0.8 | 0.20 | 2.21 | 180 | 90 |
| 7 | 1.1 | 0.15 | 1.75 | 0 | 45 |
| 8 | 1.1 | 0.15 | 1.75 | 0 | 0 |
| 9 | 1.1 | 0.20 | 2.21 | 0 | -45 |

The horizontal rotation was observed in some cases, as shown in Figure 8, where it started in 4-9 seconds and was stabilized in 25-40 seconds. The rate of rotation and stability of the ROV is varied with respect to the conditions of wave and submerged depth.

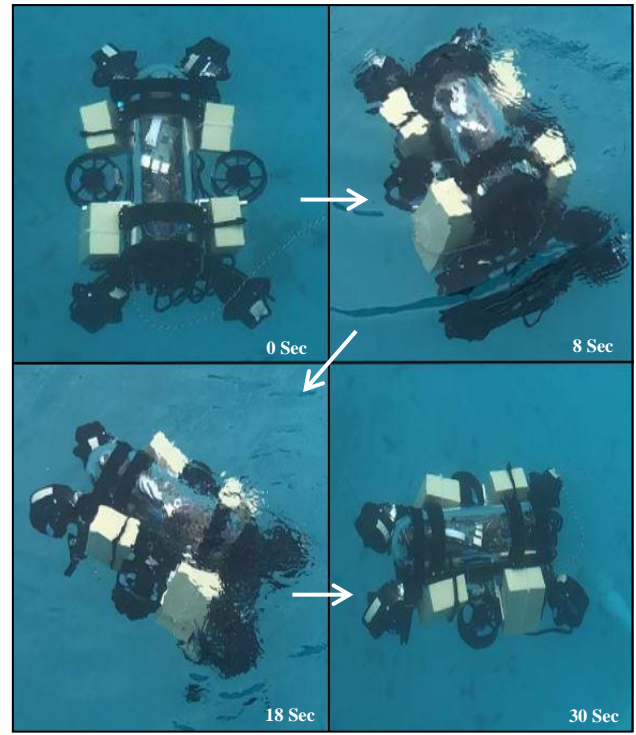

Figure 8. Horizontal rotation of ROV induced by waves.

Figure 9 compares the real-time sensor output concerning the time. Based on the navigation sensors, it was observed that as soon as the ROV depth increases, the absolute pressure is gradually growing. As a result, the pressure difference is also increased. The temperature of the water remains constant during this time.

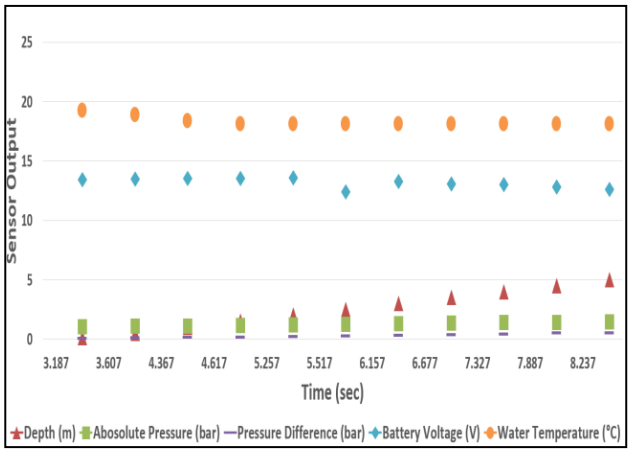

Figure 9. Real-time sensor output vs. time.

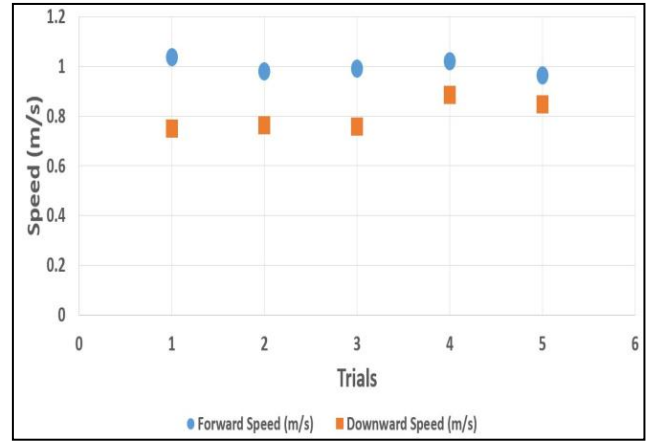

Figure 10. ROV Speed Comparison.

To calculate the maximum speed of the ROV, we need to have some hydrodynamic factors that relate to ROV and tether. The drag (D) acting on the ROV is calculated by,

$$D = T = \frac{1}{2} C_D A \rho V^2 \qquad (3)$$

Where $T$ is thrust i.e., 138.64 N, $C_D$ is the non-dimensional drag i.e., 1, $A$ is the frontal area of the ROV i.e., 0.2365473 m$^2$, $\rho$ is the density of the fresh water i.e., 998 kg/m$^3$ and $V$ is the maximum speed of the ROV, which is calculated by,

$$V^2 = \frac{2T}{C_D A \rho} \qquad (4)$$

$V$ = 1.08 m/s = 2.1 knots

Figure 10 below compares the speed-accuracy with forward and downward movement of the ROV. To calculate the forward speed of the vehicle, we used a measuring tape to measure a distance with a timer to calculate time. For downward speed, we used depth sensor data with respect to time. Based on this, to calculate the speed of the vehicle,

*Speed of the vehicle = distance traveled / time* (5)

During forward operations, we achieved the desired speed during trials 1 and 4, but with downward operations, we will not. Possible reasons can be attributed to the less number of vertical thrusters as compared to horizontal thrusters, estimated drag coefficient being lower than actual and reading errors of the depth sensor. We had an average speed of 0.99 m/s during forward operation and 0.80 m/s during downward operation.

Vehicle drag force is a vital factor that affects ROV performance. The ROV thruster needs to produce enough thrust to overcome the drag acting on the vehicle [Craven et al., 1998]. Figure 11 represents the change of total drag force with respect to the velocity of the ROV. As the velocity of the vehicle increase, the total drag force acting on the ROV is also increased in the proportion of velocity squared.

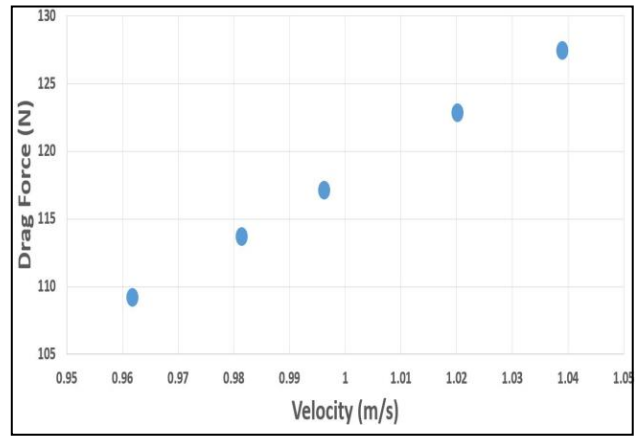

Figure 11. Change of Total Drag Force vs. Velocity.

To obtain pressure, depth, temperature, and altitude data of the ROV, a specially developed marine sensor were used, as well as the current and voltage sensing module was used to provide the battery status. For the surface navigation of the ROV, LS20031 GPS receiver is used. After doing calibrations, LS20031 requires approximately 3 minutes to give the exact location of the ROV at the surface water. The MPU-9250 9-axis sensor provides the position and orientation of the ROV in underwater. Figure 10 describes the output results for the 9-axis sensor, GPS and camera on the surface station.

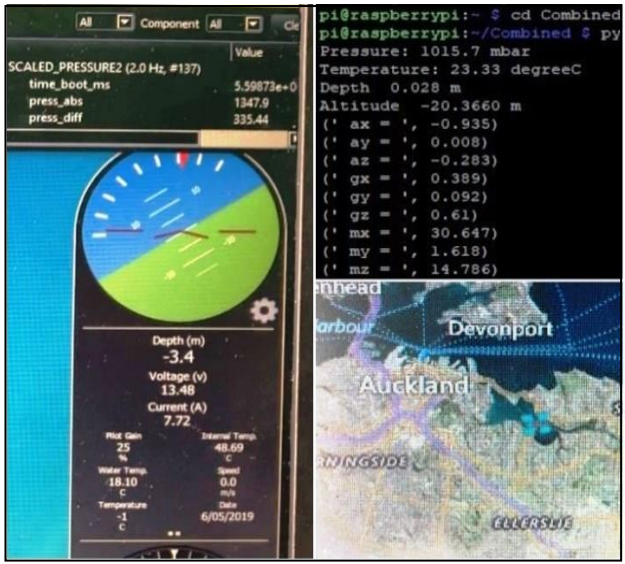

Figure 10. Output results for 9-Axis sensor, GPS and camera.

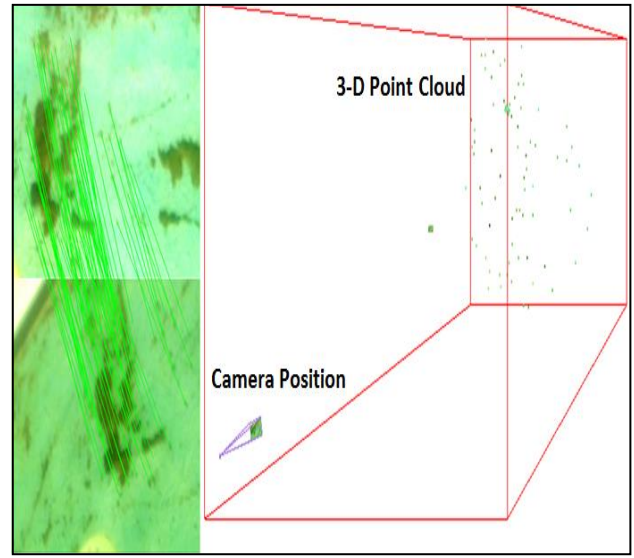

Figure 12. Pool inspection using visual SFM.

The developed ROV prototype is in its early stages of development. From the trial tests, it was seen that the ROV was able to perform all maneuvering tests like moving forward, reverse, up, down, float and submerge. The lithium-ion battery with a power supply of 14.8 V, 18 A the ROV, was able to operate underwater for approximately 3 hours. The developed ROV is able to stabilize itself in pool water without any external interference. The experimental results show that, by using the output of the Raspberry pi camera at the surface station, we can carry out the underwater marine inspection task for ship hull and marine debris inside the shallow water, as shown in Figure 11.

The video is first converted into the number of images and then get processed. Initial results show that the 3D sparse reconstruction is quite possible with a raspberry pi camera and can get a camera position with some 3D points in the cloud, but it will not work for the 3D dense reconstruction.

## 5 Conclusion and Future Work

The developed prototype of ROV has a fixed mechanical system and a modular electronic system. This paper details the design, development and qualification of the Remotely Operated Vehicle developed for carrying out the underwater marine inspections of a ship hull or marine debris etc. The initial trials of the ROV in a pool proved that the vehicle meets the primary goal of this paper and the suitability to operate in the desired marine environment.

The future work in this research includes in-detail evaluation of the developed ROV for more working depth with higher performance. Also, to make improvements in the underwater inspection task, we are going to use a high-resolution camera and the various computer vision-based algorithms. The final developed navigation system for the inspection purpose will be implemented on the BoxFish ROV. The initial test will be carried out by using developed simulation and later in a real environment.

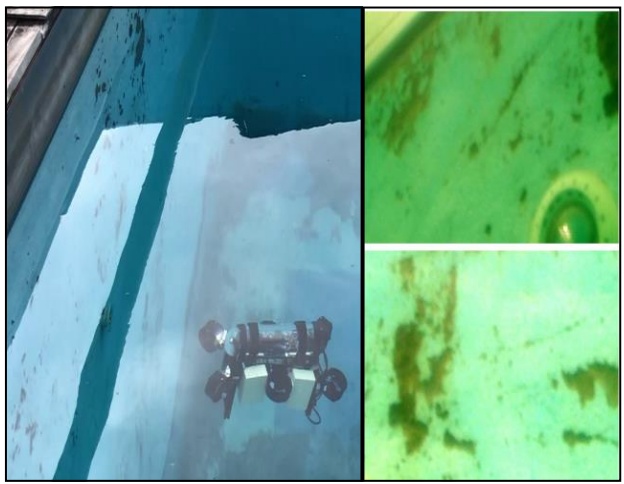

Figure 11. Pool wall inspection using the developed ROV.

To check the effectiveness and ability of the raspberry pi camera to perform the real-time required task, we further tried to process the acquired video by using a computer vision-based algorithm like visual Structure From Motion (SFM) [Marsh et al., 2013] as shown in Figure 12.

### Acknowledgments

This research work was supported by the New Zealand product accelerator (NZPA) Programme under Grant UOAX1309, funded by the Ministry of Business, Innovation and Employment (MBIE), New Zealand. The authors also want to acknowledge the support from BoxFish Research Company. We also thank Prof. Richard Green, University of Canterbury, New Zealand and Prof. Sadao Kawamura, Ritsumeikan University, Japan for their valuable guidance

.